\documentclass{article}
\usepackage{kr}

\usepackage{times}
\usepackage{soul}
\usepackage{url}
\usepackage[hidelinks]{hyperref}
\usepackage[utf8]{inputenc}
\usepackage[small]{caption}
\usepackage{graphicx}
\usepackage{amsmath}
\usepackage{amsthm}
\usepackage{booktabs}
\usepackage{algorithm}
\usepackage{algpseudocode}
\usepackage{subfigure}
\usepackage{graphicx}
\usepackage{enumitem}
\usepackage{xcolor}
\usepackage{float}
\usepackage{amsmath}
\usepackage{amssymb}
\usepackage{varwidth}

\newcommand{\GBQL}{\textsc{GBQL}}
\newcommand{\RBFQ}{\textsc{RBFQ}}

\usepackage{xcolor}
\usepackage{flushend}

\usepackage{multicol}
\title{Fitted Q-Learning for Relational Domains}

\author{%
Srijita Das$^1$\and
Sriraam Natarajan$^1$\and
Kaushik Roy$^1$\and
Ronald Parr$^2$\and
Kristian Kersting$^3$\\
\affiliations
$^1$Department of Computer Science, The University of Texas at Dallas, USA\\
$^2$Department of Computer Science, Duke University, USA\\
$^3$Department of Computer Science and Centre for Cognitive Science, TU Darmstadt, Germany\\
\emails
\{Srijita.Das, Sriraam.Natarajan, Kaushik.Roy\}@utdallas.edu,\\
parr@cs.duke.edu, kersting@cs.tu-darmstadt.de
}

\begin{document}

\maketitle

\begin{abstract}
We consider the problem of Approximate Dynamic Programming in relational domains. Inspired by the success of fitted Q-learning methods in propositional settings, we develop the first relational fitted Q-learning algorithms by representing the value function and Bellman residuals. When we fit the Q-functions, we show how the two steps of Bellman operator; application and projection steps can be performed using a gradient-boosting technique. Our proposed framework performs reasonably well on standard domains without using domain models and using fewer training trajectories.
\end{abstract}

\section{Introduction}
Value function approximation in Reinforcement Learning (RL) has long been viewed using the lens of feature discovery~\cite{parr2007analyzing}. A set of classical approaches for this problem based on Approximate Dynamic Programming (ADP) is the fitted value iteration algorithm ~\cite{boyan1995generalization,ernstGW05,riedmiller2005neural}, a batch mode approximation scheme that employs function approximators in each iteration to represent the value estimates.
Another popular class of methods that address this problem is  Bellman error based methods~\cite{menache2005basis,keller2006automatic,parr2007analyzing}. The key intuition is that the Bellman error has positive dot product with the true value function and, thus, adding basis functions based upon the Bellman error can lead to a good approximation to the true value function. 

Here, we consider relational domains that are typically described using parameterized state-action spaces. 
While it is conceivable to instantiate each object and construct a grounded MDP, given the variable number of objects, this can yield substantially large state-action spaces that render solving the grounded MDP intractable in practice. On the other hand, as noted by Tadepalli et al.~\shortcite{tadepalli2004relational}, typical function approximators do not generalize well when applied to relational domains. 
Consequently, a class of RL methods, called Relational Markov Decision Process (RMDPs), have been developed for learning optimal behaviour in these worlds, and methods that directly learn and reason at the first-order level have been developed in the broad area of Relational Reinforcement Learning (RRL)~\cite{price2001imitation,sanner2009practical,wang2008first,tadepalli2004relational,wu2007discovering}. While specific methodologies differ, most of these methods operate at a symbolic level and define the mathematical operations on these symbols to learn the values of the parameterized states (which can essentially be viewed as a ``group" of states in classical RL). Most of these methods, however, are exact methods. A notable exception is by Guestrin et al.~\shortcite{guestrin2003generalizing}, who developed approximate learning for RMDPs by representing and learning value functions in a stage-wise manner. 

Inspired by the success of approximate value function learning for propositional domains, we propose the {\bf first set of approximate Q-value function learning methods} for relational domains. We take two specific approaches -- first is to represent the {\bf lifted} Q-value functions and the second is to represent the Bellman residuals -- both using a set of relational regression trees (RRTs)~\cite{blockeel1998top}. A key aspect of our approach is that it is {\bf model-free}, which most of the RMDP algorithms assume. The only exception is Fern et al.~\shortcite{fern2006approximate}, who directly learn in policy space. Our work differs from their work in that we directly learn value functions and eventually policies from them and adapt the most recently successful relational gradient-boosting (RFGB)~\cite{natarajan2014introduction}, which has been shown to outperform learning relational rules one by one.


Our work can be seen as learning functions over relational domains that permit the efficient combination of these functions without explicitly considering the entire space of models. 
Wu and Givan~\shortcite{wu2007discovering} approximate the value function by using beam search with relational features that are iteratively learned from sampled trajectories. Our approach uses gradient boosting where we learn sets of conjunctive features as paths from the root to the leaf of each RRT and lift the arguments of fitted Q-learning in propositional domains~\cite{ernstGW05,tosattoPDR17} to relational domains. Indeed, if we knew that the target Q function belongs to a specific class and use this information to model the value function there is no need of trees and boosting. However, in relational domains this information is almost never available: {\em without difficult feature engineering, the shape of the value function is almost always unknown} -- a reason why some work directly operated at the policy level~\cite{fern2006approximate,kersting2008non}. 

More precisely, we make the following key contributions: (1) We develop a unified framework for handling relational RL problems by exploring the use of relational trees to model the Q-values and Bellman residuals. (2) We outline the connections between the boosted approach to the classical RL methods for value function approximation (Bellman error, aggregate and tile coding methods). We show that our method can be viewed as encompassing these different approximation techniques. (3) Finally, we demonstrate empirically the effectiveness and the generalization ability of our proposed approach where our results are {\em on par with other RMDP/planning methods without using domain models and with fewer training trajectories}. Without extensive feature engineering, it is difficult---if not impossible---to apply standard methods within relational domains in which there is a varying number of objects and relations. 

The rest of the paper is organized as follows. After introducing the necessary background on RL, value function approximation and relational learning in the next section, we outline the algorithm and analyze its properties. Before concluding and presenting future work, we present the empirical evaluation of the proposed approach on three classical Relational RL domains. 

\begin{figure}[t]
\centering
\frame{\includegraphics[width=0.47\textwidth]{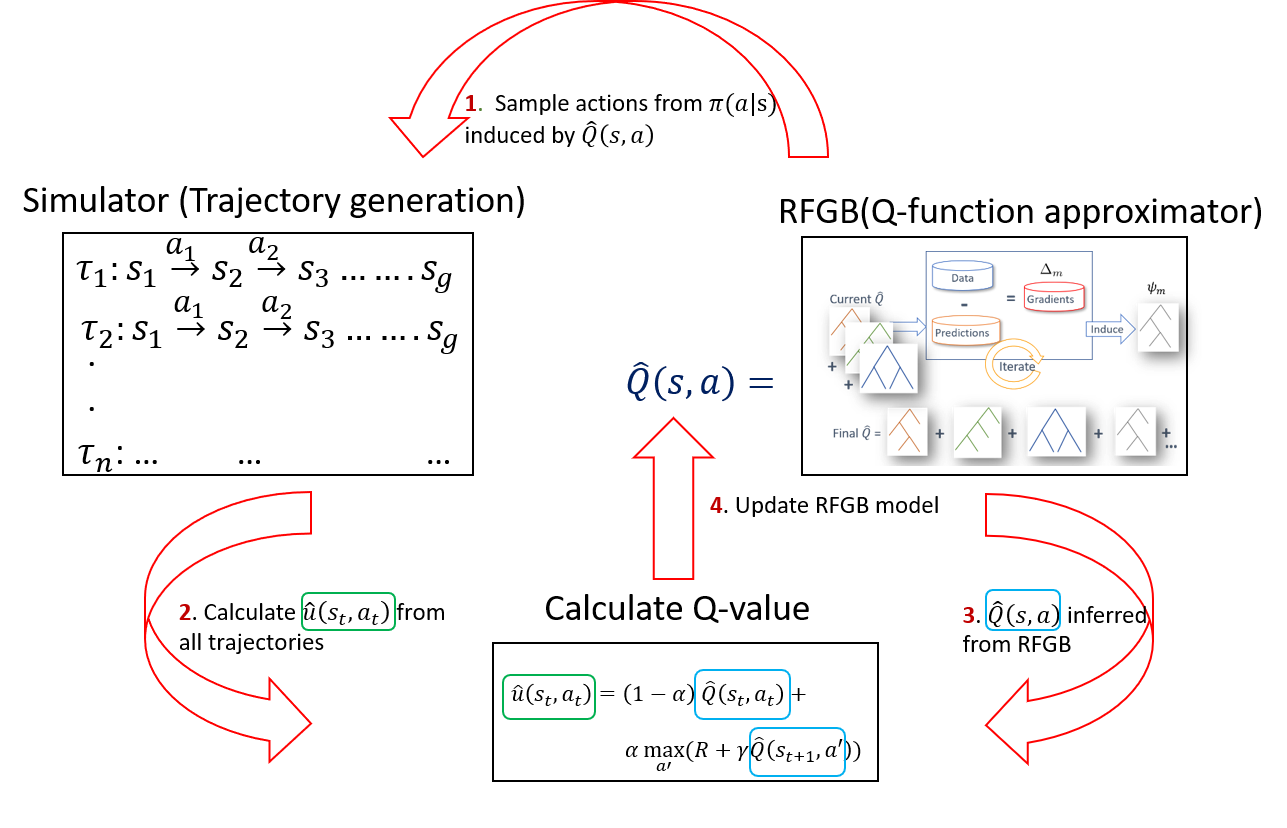}}
\caption{Proposed framework for GBQL}
\label{RRL_framework}
\end{figure}

\section{Background}
\paragraph{Markov Decision Processes (MDPs):} 
An MDP is described by a tuple $(\mathcal{S},\mathcal{A},\mathcal{R},P,\gamma) $ where $\mathcal{S}$ is the state space, $\mathcal{A}$ is the action space, $\mathcal{R}$ is the reward function,$P$ is the transition probability defined as $P:\mathcal{S}\times \mathcal{A} \rightarrow [0,1] $. For infinite horizon problems, a discount factor $\gamma \in [0,1)$ is specified to trade-off between current and future reward. In state $s_t$, after taking action $a_t$, the agent receives a reward $R_t \sim \mathcal{R}$. A policy $\pi$ is a mapping from $\mathcal{S}$ to a probability distribution over action-space $\mathcal{A}$ given a state. The optimal Q-value (expected reward) of a particular state action pair is given by the Bellman optimality equation:
\begin{equation}
    Q^*(s,a) = \sum_{s'}P(s'|s,a)[ R + \gamma\max_{a'} Q^* (s',a')]
\end{equation}
\textbf{Bellman Error:} For a fixed policy $\pi$, the true Q function $Q^\pi$ can be complex and might not lie in the subspace of representable value functions considered by the hypothesis class of a function approximator. A projection operator $\Pi$ is used to project the true value function to the representable subspace of function represented by the approximator. The Bellman error is the expected difference between value estimates from successor states and current state: $BE(s,a)=$
\begin{align}
   \left(\sum_{s'}P(s'|s,a)[R+\gamma \max_{a'} Q(s',a')]\right)-Q(s,a) \label{eqn:be}
\end{align}
The estimate $Q(s,a)$, which is an approximation of the true function, is the projected parameterised Q function in the representable function space. The empirical Bellman operator $T^*:\mathbb{R}^{|\mathcal{S}|\times |\mathcal{A}|} \rightarrow \mathbb{R}^{|\mathcal{S}| \times |\mathcal{A}|}$ is the first term in the above equation with sampled next states rather than an explicit expectation. The empirical Bellman error can be defined as $T^*Q(s,a)-Q(s,a)$. Applying the Bellman operator to a Q-function in a representable subspace may result in a Q function in a different subspace, necessitating projecting back to the representable subspace using a projection operator $\Pi$.



\textbf{Approximate Dynamic Programming (ADP):} 
For large domains, approximate methods can be used to estimate the Q-function so that the function need not be stored for every state-action pair. Some of the popular approaches for Q-function (or value function) approximation in propositional domains like Least Squares Temporal Difference (LSTD) Learning~\cite{boyan1999least}, Least Squares Policy Evaluation (LSPE)~\cite{yu2009convergence} and Least Sqaure Policy iteration (LSPI)~\cite{lagoudakis2003least} approximate the Q-function as linear function of the parameter vector i.e: $\hat{Q}(s,a) = \mathbf {w}\cdot \mathbf{\phi}$ where $\mathbf{w}$ is the parameter of the Q-function and $\mathbf{\phi}$ is the feature vector representing the state space. These features are also called {\em basis functions} and are ideally linearly independent of each other.  Our work can be seen as extending this line of work by \emph{ approximating the Q-functions and/or the Bellman residuals using a non-parameteric relational representations}.

The ADP framework is a collection of RL methods where the values or policies are approximated using information from samples. LSTD, for example, uses a least squares approximation to the temporal difference learning paradigm. 
The main component of ADP approaches such as LSTD is the computation of weights for the approximation of the value function. 
In relational domains, groups of states are typically indistinguishable w.r.t. the values that they assume allow {\em parameter sharing}. For example, in a blocks world, many configurations could have similar ground state values as they exhibit similar properties such as similar towers etc. Consequently, it is possible to learn a {\em lifted representation} such as the one by Wu and Givan~\cite{wu2007discovering} that can reason over sets of feature values.

\textbf{Fitted-Q learning:}
Fitted Q-learning~\cite{ernstGW05} is a form of ADP which approximates the Q-function by breaking down the problem into a series of regression tasks. Hence, each iteration of Fitted-Q iteration is similar to essentially solving a supervised regression problem.
At every iteration of Fitted-Q iteration, the empirical Bellman operator is applied to the current Q-function approximation. Each Regression model then tries to find the best parameters to minimize the regression error. The Q-functions learned by such regression models lie in a subspace of all functions represented by the regression model and hence are an approximation of the true function.

We employ Functional Gradient Boosting~\cite{friedman2001greedy} in relational domains~\cite{getoor2007introduction,DeRaedt2016} to capture the Bellman error for every Q-learning iteration and obtain {\em lifted} basis functions in a non-parametric fashion. 
In domains where we require probabilistic policies, a Gibbs distribution over the values can be used to model probabilistic action choices. While it is conceivable to learn policies directly as performed by policy gradient approach of Kersting and Driessens~\shortcite{kersting2008non} or imitation learning by Natarajan et al.~\shortcite{natarajan2011imitation}, our methods approximate the Q-values or the Bellman error using gradient boosting.


\textbf{Tree-based RRL:} Our work is inspired by Dzeroski et al.~\shortcite{dvzeroski2001relational} on learning single regression tree for RRL. This work was later extended by others~\cite{driessens2003relational,driessens2004integrating,driessens2001speeding,gartner2003graph} w.r.t building incremental RRTs, providing guidance to the learner and regression algorithms using kernels and gaussian processes. However, our work is different in two key aspects: (1) We are more sample efficient in that we do learn only from a smaller set of sampled trajectories in each iteration while their methods use all the trajectories when learning the Q function, and (2) we adapt multiple stage wise grown boosted RRTs for Q function approximation in two different ways as discussed later.

\section{Gradient Boosted Fitted Q-learning}

{\bf Set up:} We consider learning to act in {\em relational, discrete, noisy} domains where the domains consist of multiple interacting objects. We adopt the RMDP definition of Fern et al.~\cite{fern2006approximate}, where $\mathcal{S}$ and $\mathcal{A}$ of the original MDPs are represented by a set of objects $\mathcal{O}$, a set of predicates $\mathcal{P}$ and set of action types $\mathcal{Y}$. As with their definition, we use {\em facts} to denote a predicate grounded with a specific instantiation. For example $on(a,b)$ is a grounding of a predicate $on(X,Y)$. Thus a set of ground facts will then define the state of the MDP and  $\mathcal{S}$  is the set of all the possible facts in the domain. We make a closed-world assumption, in that facts that are not observed are assumed to be false. Similarly, the action is a grounded action type. For example $pickup(a)$ is an action which is obtained by grounding $pickup(A)$ with $A=a$. 

Given this problem definition, the goal is to learn $Q(s,a), \forall s,a$. However, the learning occurs at a symbolic representation, i.e., we learn $Q(X,A)$ for all the objects (or partial instantiations of the objects $X$) and the action types $A$. Consequently, the learned representation is a lifted one while the training examples/observations are at the propositional level. To represent these lifted representations, as mentioned earlier, we use RFGB~\cite{natarajan2014introduction,natarajan2012gradient,natarajan2011imitation} and a single tree, TILDE~\cite{blockeel1998top}, for approximating the Q-value functions. 

The key idea in RFGB is that like its propositional counterpart, a functional representation is used to model the Q-function values. The gradients are obtained w.r.t. this function and are approximated using regression functions at each step of gradient computation. In relational domains, these regression functions are {\em Relational Regression Trees} (RRTs)~\cite{blockeel1998top}. In this work, we consider three types of Q-value function approximations -- single relational tree, and two variants of boosted relational trees.

 \textbf{Boosted Regression Trees for Gradients:}
During learning, in each iteration of RFGB, gradients are computed for each training example w.r.t the loss function and an RRT is learned at each gradient step to fit to these gradients from previous iterations. The intuition is that the direction of these approximately learned gradients is similar to the original gradients. Given a set of positive examples and potentially some negative examples (in some cases, these are computed using closed-world assumption), the goal is to fit a function $F$ as a set of relational regression trees. 
While previously this idea had been employed in the context of relational classification~\cite{natarajan2012gradient,khot2011learning,ramanan2018structure}, the use of RFGB for value function approximation, requires extending them to regression. 

In the context of regression, a typical loss function used is the squared error. The squared error with respect to a single example $(x_k,y_k)$ is $l(x_k,y_k) = (y_k-F(x_k))^2$, where $y_k$ is the true (regression) value and $F(x_k)$ is the value estimated by the regression function $F$. The (functional) gradient of an example $x_k$ at the end of m iterations is, 
$\Delta_m^k$ $=$ $\frac{\partial l(x_k,y_k)}{\partial F_{:m-1}(x_k)}=$ ${y_{k} - F_{:m-1}(x_{k})}$, 
 where $F_{:m-1}$ denotes the sum of all the gradients computed through the $m^{th}$ iteration of relational functional gradient boosting. Since we are in a functional space, the gradients are calculated with respect to the function $F$ and not the parameters. The final regression function $F$ after $m$ rounds of boosting is $ F_m=F_0+\Delta_1+\Delta_2+\cdots+\Delta_m$.

In our case, $y_k$ is the Q value of the current state-action pair ($s_k,a_k$) \footnote{We use $i$ to denote the Q-learning iterations, $k$ to denote the example index (state-action pairs) and $m$ to denote the boosting iterations from here on.} after applying the empirical Bellman operator $T^*$ and $F$ is the Q-value for the current state-action pair as approximated by the model $F_m$. $x_k$ in the above equation corresponds to all the relevant features of the current state $s_k$. Recall that in relational domains, for the same set of predicates, different states could have different number of instantiated (grounded) features. For instance, the predicate $on(X,table)$ can yield different number of blocks in different states. RFGB employs RRTs to lift the representation such that it is independent of the number of ground instances. 
To account for this, we redefine the loss function as,

\begin{equation}
    l(s,a)=\sum_{(s,a) \in \tau}(\hat u(s,a) -\hat Q(s,a))^2 \label{sqloss}
\end{equation} 
where $\hat{u}(s,a)$ is the point wise regression value which gets fitted to the new model. 
$\tau$ refers to the set of sampled training trajectories.

\begin{algorithm}[t]
\begin{algorithmic}[1]
\Function{$\mathtt{\GBQL}$}{$N$,$M$,$p$,$\rho$}
\State $\mathbf{Initialize:}$ Set $\hat{Q}_0$ := 0
\For {$1 \leq i \leq N$} \Comment{$\#$ of Q-learning iterations}
\State $\mathbf{E} := \{\varnothing\}$ \Comment{Stores tuple $(s_k,a_k,\hat{u}(s_k,a_k))$}
\For {$1 \leq j \leq p$} 
\State \Comment{generate mini batch of trajectories}
\State $s_0\sim P(\mathbf{S})$ 
\State \Comment{Choose $s_0$ from initial state distribution}
\State Generate $\tau_j$ starting from $s_0$ by accessing  $\rho$
\EndFor
\For {$1 \leq k \leq |\mathbf{D}|$} 
\State \Comment{Iterate over all $(s_k,a_k,s_{k+1})$} in $\tau$
\State \begin{varwidth}[t]{\linewidth}
$\hat u(s_k,a_k)$ := $(1-\alpha)\hat{Q}_{:i-1}(s_k,a_k)$\par
\hskip\algorithmicindent $+\alpha[R+\gamma \max_{a'}\hat{Q}_{:i-1}(s_{k+1},a')]$
\end{varwidth}
\State $\mathbf{E}=\mathbf{E}\cup\{s_k,a_k,\hat u(s_k,a_k)\}$
\EndFor
\State $\hat{Q}_i = \Call{TreeBoost}{\mathbf{E},M}$
\EndFor
\State \Return $\hat{Q}_N$
\EndFunction
\end{algorithmic}
\caption{$\GBQL$ learning}
\label{algo:gbql}
\end{algorithm}

Thus, the goal is to minimize the difference between the regression value of the current state-action pair (empirical value) and the current Q-value according to the model. For every {\bf lifted state-action pair} $(s_k,a_k)$,  $\hat u(s_k,a_k)$ referred in~\eqref{sqloss} is obtained by applying the Bellman operator,
\begin{equation}
\begin{array}{cl}
\hat u(s_k,a_k) & = (1-\alpha)\hat{Q}_{:i-1}(s_k,a_k) \\
& + \, \alpha[R + \gamma \max_{a'} \hat{Q}_{:i-1}(s_{k+1},a')]
\end{array}
\end{equation}
$\alpha$ is a learning rate and $\hat Q_{:i-1}(s_k,a_k)$ is the Q-value of ($s_k,a_k$) as returned by RFGB through the $(i-1)^{st}$ Q-iteration, $s_{k+1}$ refers to the successor state of $s_k$ in the sample. The learning rate $\alpha$ is set close to 1 so that the previous estimates of the current state-action values have a small contribution to the new estimate.
At each boosting iteration $m$, every training example $k$ includes a gradient $\Delta_{i:m}^k$ and a new RRT is trained to fit these $\Delta_{i:m}^k, \forall i$.  The gradient $\Delta_{i:m}^k$ is computed as,
\begin{equation}
    \Delta_{i:m}^k=\hat{u}(s_k,a_k)-\hat{Q}_{m-1:i}(s_k,a_k) \label{eqn_gradient},
\end{equation}
where
\begin{equation}
\hat{Q}_{m-1:i}(s_k,a_k) = \Delta_{i:1}^k + \Delta_{i:2}^k + \ldots + \Delta_{i:m-1}^k
\end{equation}
$\hat{Q}_{m-1:i}$ is the Q value after $i$ iterations of Q-learning and $m-1$ iterations of boosting. The final regression value over a lifted state-action pair $(s_{k},a_{k})$ is the sum of regression values after all $M$ rounds of boosting: $ \hat{Q}_{:i}(s_k,a_k) = $
\begin{equation}
   \hat{Q}_{M:i}(s_k,a_k) = \Delta_{i:1}^k + \Delta_{i:2}^k + \ldots + \Delta_{i:M}^k \label{eqn:final_q}
\end{equation}
The final Q-function, after $N$ Q-iterations, is $\hat{Q}_{:N}$, and it defines a value for each lifted state as a sum of $M$ gradients.

\begin{algorithm}[t]
\begin{algorithmic}[1]
\Function{TreeBoost}{$E$,$M$}
\State $\mathbf{Initialize:}$ Set $F_0$:=\Call{LearnRRT}{$E$} 
\State \Comment{Learn relational regression tree to fit $\hat{u}(s,a)$}
\For {$1 \leq m < M$} \Comment{Iterate through $M$ gradient steps}
\State $F_m:=F_{m-1}$
\State $\S_i$=\Call{Gengradients}{$E$,$F_m$,$\mathcal{A}$}\Comment {Compute gradients, for action $a_i\in\mathcal{A}$}
\State $\Delta_m$=\Call{LearnRRT}{$E$,$\S_i$} \Comment{Learn relational regression tree to fit the gradients $\S_i$}
\State $F_m:=F_m+\Delta_m$
\EndFor
\State \Return $F_M$
\EndFunction
\end{algorithmic}
\caption{TreeBoost learning}
\label{algo:gbtree}
\end{algorithm}

\begin{figure}[b]
\caption{An example \emph{Relational Regression Tree} representing the structure of the Q-function for \textit{Unload} action on Logistics domain.\label{rrt}}
\includegraphics[width=0.9\columnwidth,scale=0.5]{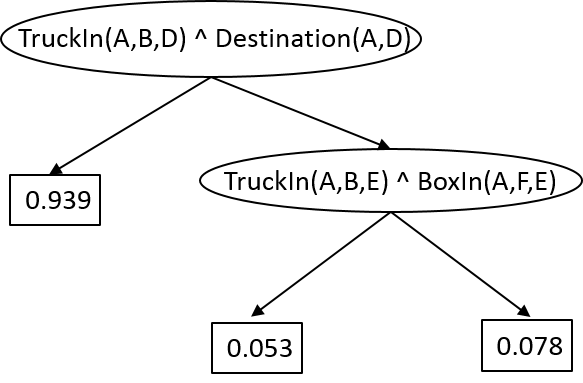}
\end{figure}

An example of a RRT learned in the logistics domain is shown in Fig.~\ref{rrt}, where the inner nodes represent a first order conjunction (test). The left branch indicates that the test is satisfied and the right branch is when it does not hold. The root node checks if a truck $B$ is in the destination city $D$ at the current state. This state will have the highest Q-value. The leaf values denote the Q value of the state-action pair that satisfy any particular path. Note that all the states and actions in this Figure are {\bf lifted} i.e., parameterized. In the general case, these could be partially grounded, for instance {\em TruckIn(A,Paris,E)}. RFGB can learn at the fully lifted level, partially instantiated level or the ground level based on the language bias provided during learning.

With RRTs for computing gradients at hand, we now present the algorithm for learning the approximate Q-functions of relational MDPs using RFGB. Specifically, Algs~\ref{algo:gbql} and ~\ref{algo:gbtree} present the outline of our proposed approach, called Relational \underline{G}radient-\underline{B}oosted \underline{Q}- \underline{L}earning ($\GBQL$). $\GBQL$ takes as input the number of Q-learning iterations \textit{N}, the number of iterations for basis function computation $\textit{M}$, the number of training trajectories for each iteration of Q-learning $\textit{p}$ and an access to the domain simulator $\rho$ (a setting similar to that of Parr et al.~\shortcite{parr2007analyzing}). At each iteration a set of trajectories $\tau$ are sampled from the simulator $\rho$. The relational simulator $\rho$ takes as input the current state $s_t$ and action $a_t$ at time step t and returns the next relational state $s_{t+1}$. For every $(s_t,a_t)$ pair, the simulator also returns a reward function $R_{t+1}$ which is designed according to the problem being solved. At every Q-learning iteration, $p$ mini batches of trajectory $\tau_1,\tau_2,\cdots \tau_p$ are sampled from the simulator. A relational trajectory $\tau_i$ is defined by a sequence of relational facts and actions taken over a sequence of time steps until a fixed predefined goal is reached, i.e $\tau_i=\{(s_0,a_0),(s_1,a_1),\cdots(s_g,.)\}$ where $s_g$ is the goal state. The initial state $s_0$ is chosen from an initial state distribution $P(\mathbf{S})$. For every $\langle state, action, next state\rangle$ tuple in the trajectory ( such tuples in every iteration is appended to a set $\mathbf{D}$) , the value for the current state-action pair is updated using the Bellman operator (line $13$). 
Next, every state-action pair and its corresponding q-value after applying Bellman operator are added to the training set for a function call to the \emph{TREEBOOST} algorithm (lines $14-16$).

For every $\langle state, action, value \rangle$ tuple in the data set $\mathbf{E}$, the goal is to learn a set of RRTs to approximate the Q-values compactly.
In our case, these correspond to finding the combinations of the set of (relational) features that best approximate the value of the current state-action pair. Note that in each iteration of the \emph{TREEBOOST} procedure (indexed by $m$), a single regression tree is learned and added to the initial model (line 2). The \emph{LEARNRRT} function takes the examples and the initial q-values $\hat{u}(s,a)$ as input and learns a single RRT. 
An RRT is learnt over relational features by scoring each test node (we use weighted variance as scoring function) and choosing the best node to split. In the next boosting iteration, the regression value is updated with the difference between the original value $\hat{u}(s,a)$ and the value returned by the current set of trees as shown in~\eqref{eqn_gradient} -- function call to \emph{GENGRADIENTS} in line 6. The key here is that \emph{TREEBOOST} is essentially performing a series of gradient steps to best fit the Q value of each state-action pair according to the sampled training set. Each tree corresponds to a single gradient in that direction, and the final value is the sum of all the values from the different regression trees as shown in~\eqref{eqn:final_q}. Hence, each call to \emph{TREEBOOST} returns single set of regression trees for every lifted action type in $\mathcal{A}$.

Now, in the main $\GBQL$ procedure, the set of trees (Q-function) learned in the previous iteration are used to update the Q value of state action pair at the current iteration while applying the Bellman operator. To generate the next set of trajectories for the next iteration from the simulator $\rho$, we follow an $\epsilon$-greedy policy where we choose a random action with a probability $\epsilon$ and the best action with respect to the current Q-function estimates with probability $(1-\epsilon)$ for a given state. Figure~\ref{RRL_framework} shows how the simulator $\rho$ is integrated with the $\GBQL$ framework.

\begin{figure*}[t]
\centering
\includegraphics[width=\textwidth]{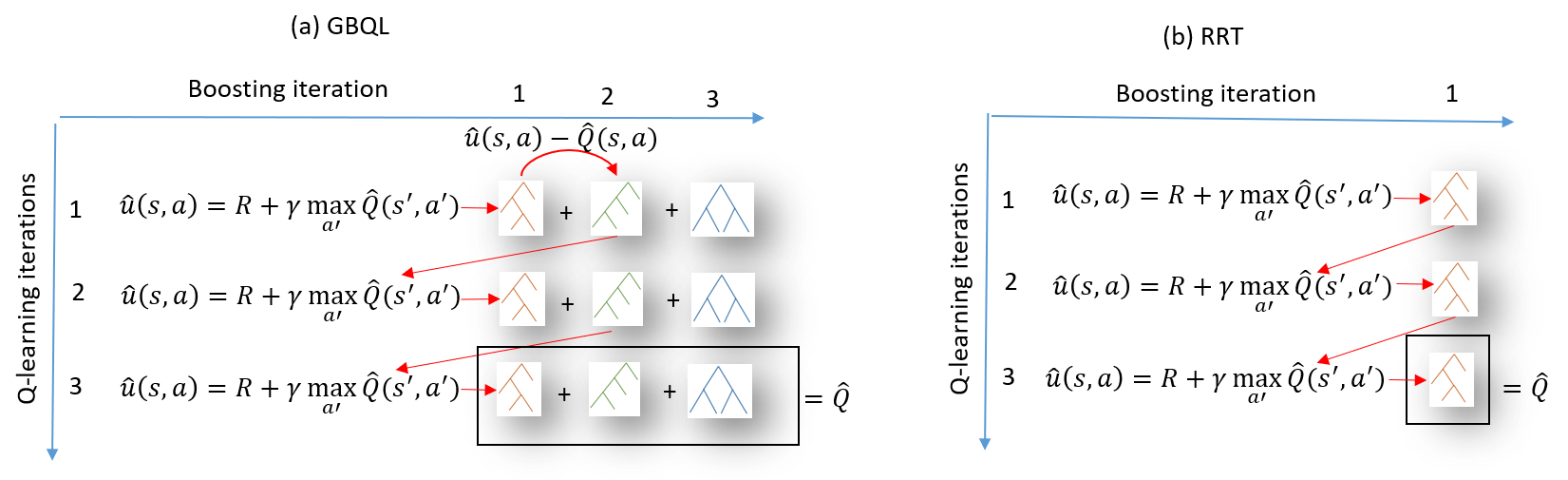}
\caption{Boosted \textbf{GBQL} and non-boosted (\textbf{RRT}) Q-function representations for fitting Bellman error.\label{gbql+rrt}}
\end{figure*}

These set of learned regression trees at any iteration are the (relational) basis functions for that iteration. One could also not boost the trees but directly fit a single RRT to the q-values. These form the two different methods for fitted-q as shown in Fig.~\ref{gbql+rrt}.

\begin{figure}[t]
\includegraphics[width=0.45\textwidth]{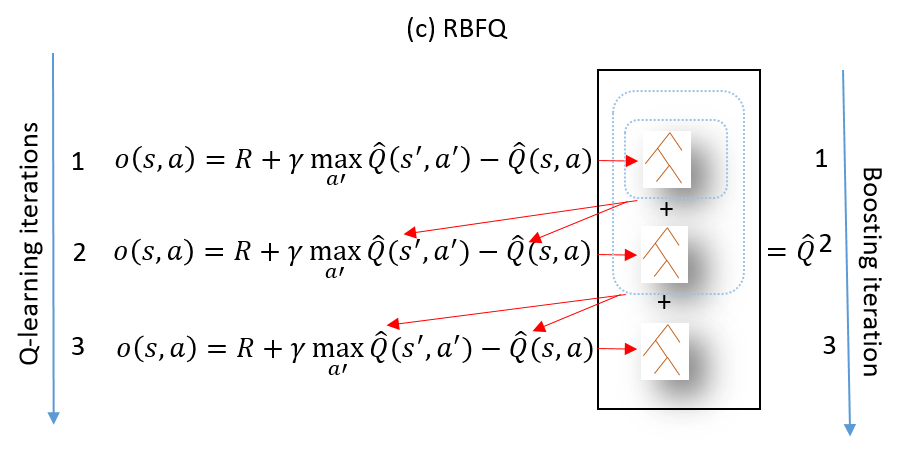}
\caption{Boosted \textbf{RBFQ} Q-function representations for fitting Bellman error.\label{rbfq}}
\end{figure}

\textbf{Single Regression Tree for Bellman Residuals:}
We adapt the work of Tosatto et al.~\shortcite{tosattoPDR17} for relational domains and present an alternate method of boosting. Fig.~\ref{rbfq} presents an overview of our proposed approach, which we call \underline{R}elational \underline{B}oosted \underline{F}itted-\underline{Q} ($\RBFQ$). We explicitly calculate the \textit{Bellman residual} $o(s_k,a_k)$ from samples in every iteration $i$ of Q-learning as, $ o(s_k,a_k)=$
\begin{equation}
   [R+\gamma \max_{a'}\hat{Q}_{:i-1}(s_{k+1},a')]-\hat{Q}_{:i-1}(s_k,a_k) \label{eqn:bellman_error}
\end{equation}
where $\hat Q_{:i-1}(s_k,a_k)$ is the Q-value of ($s_k,a_k$) as returned by $\RBFQ$ through the $(i-1)^{st}$ Q-iteration and is defined as 
\begin{equation}
\hat{Q}_{:i-1}(s_k,a_k)=\sum_{j=1}^{i-1}\hat{Q}_{j}(s_k,a_k) \label{eqn:additive_q}
\end{equation}
$\hat{Q}_{j}(s_k,a_k)$ is the Q-value of $(s_k,a_k)$ as returned by the $j^{th}$ Q-iteration.
$o(s_k,a_k)$ in Equation~\ref{eqn:bellman_error} is equivalent to the Bellman error estimate defined in Equation~\ref{eqn:be} except that instead of expected Bellman error over all possible successor states sampled from the underlying transition function, this is calculated directly from the sample.
We compute the Bellman residual $o(s_k,a_k)$ for each state-action pair in the trajectory set and fit a single weak RRT. The representation learnt for the Bellman residual is a lifted representation where every inner node of the RRT contains conjunction of first order logic predicate and each leaf node contains the approximated Q-value of sets of state-action pair that satisfies the test conditions along a branch. For Q-learning iteration $i$, the learnt Q-function is an additive one over all the previous Q-functions approximated so far as in Equation~\ref{eqn:additive_q}. 

Note that while $\RBFQ$ algorithm is similar in spirit to $\GBQL$, there are several key differences. While in $\GBQL$, we apply the Bellman operator $T^*$ and learn the first RRT in every Q iteration, here,we calculate the Bellman residual ($T^*Q-Q)$ and learn the RRTs. The Bellman residuals can be seen as the gradients $\Delta_m$ for $\GBQL$, however, these gradients are learnt in different Q-learning iterations unlike $\GBQL$. Another key difference is that in $\GBQL$, in every Q-learning iteration, the Bellman operator is applied once and then the gradients are fitted by a series of Gradient Boosted RRTs in the same iteration. However in $\RBFQ$, in every iteration, an RRT is learnt for Bellman residual and in subsequent Q learning iterations, an additive model over all the previously learnt Q-functions are considered for calculating the Bellman residual. We hypothesize that {\em directly fitting the Bellman residuals can lead to better fits in some important states, approximating the Q-values could result in an improvements on the entire state space on an average.} We verify this hypothesis empirically. 


\section{Relation to Classical ADP Approaches}

Aggregate approaches group (cluster) states and learn values over these clusters. For instance, let $S=\{s_1,...,s_n\}$ denote the set of states and $C= \{C_1,...,C_R\}$ the clusters. Then the value of a state is calculated as,
$V(s) = \sum_i \beta_i\cdot I(s \in C_i)$
where $\beta_i$ is the weight on the cluster. In our method, the grouping is obtained naturally. Each path from root to leaf is a first-order logic clause which is a {\em lifted} representation of the ground states. $\beta_i$ can be viewed as the weights learned in the leaves of each branch of a tree. Since each state only satisfies one path, each tree will yield only one $\beta$ for a state. A key advantage of \textsc{GBQL} over the aggregate features is that in the latter, the transition function of one state from a cluster to another $P(C_i(s)|C_j(s),a)$ is not easy to compute and need to be designed carefully, a problem that is avoided in our model-free setting.

It is possible to view our \textsc{GBQL} procedure as a projection into a space of piecewise-constant value functions, but such value functions have limited representational power. 
One way to get greater representational power would be to keep trees or entire Q-functions from previous iterations. Previous Q-functions would now become basis vectors and Q-function approximation could be used to combine these basis vectors to form a new Q-function. 
We will explore this connection in the future. Currently, we discard the trees from the previous iteration for efficiency and scalability to large relational tasks in $\GBQL$. 

Indeed, a related method for solving ADP is that of tile-coding~\cite{whiteson2007adaptive}, a linear piece-wise approximator that partitions the state space into (potentially overlapping) regions called {\em tiles}. These are essentially axis-parallel hyper-cuboids learned from the original state space. It is easy to see that our method can be considered as learning these tiles since each tree can be considered as defining a tile in the original space. While the original method~\cite{albus1981brains} also used Bellman error to identify the tiles, it was a heuristic based method. \textsc{GBQL} on the other hand, can be viewed as performing a gradient descent in the Bellman error space (when viewed in a functional form).

Another direction related to our work is the recent work on Deep RL. Deep Q networks (DQN), first introduced by Mnih et al.~\shortcite{mnih2015human} and later extended by others~\cite{hausknecht2015deep,van2016deep,mnih2016asynchronous,anschel2017averaged} can also be seen as fitted Q-iteration where the Q-functions are approximated by Deep Neural networks. Similar to DQN which uses a non-linear function approximator to estimate the Q values, we use gradient boosted trees which are also non-linear function approximators. The major difference between this line of work with ours is that while they operate at propositional level and approximate the Q-values of every state action pair encountered, we lift the Q-functions to the relational setting where the gradient boosted first order trees are used to capture the structure of the domain. Recently, Zambaldi et al.~\shortcite{zambaldi2018deep} proposed to use attention mechanism to capture the interactions between the various entities in the domain. While this work captures the relation existing in structured domain, they are implicitly captured by the attention aware Neural network architecture ~\cite{santoro2018relational} and therefore relations are not specified before learning. We on the other hand, employ {\em a symbolic representation that specifies relationships between entities through expressive first order logic predicates} explicitly. Thus we allow for encoding domain and common sense knowledge in a meaningful way. Finally,  Jiang et al.~\shortcite{zhengyao2019neural} used symbolic methods with Neural networks by using differentiable ILP which can directly feed in as a differentiable function to Neural networks. While they use policy gradient and operate in policy space, we use lifted representation in the Q-function space and non-differentiable ILP to capture symmetries existing in the state space.



\section{Experimental Evaluation}
 \begin{figure*}[!ht]
    \centering
    {
    \subfigure[Stack]{
    \includegraphics[width=0.235\textwidth,height=3cm]{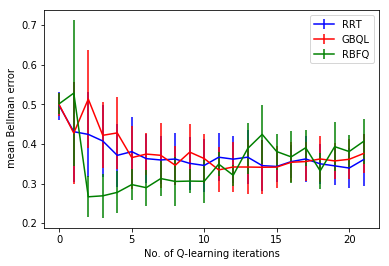}
    \label{mean_bellman_blocks_sb}
    }
    \subfigure[Unstack]{
    \includegraphics[width=0.235\textwidth,height=3cm]{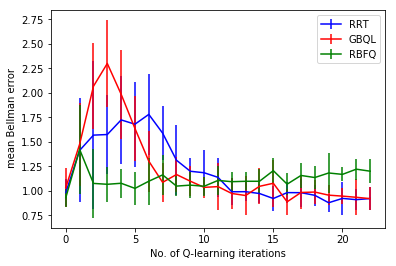}
    \label{mean_bellman_ub}
    }
    \subfigure[On]{
    \includegraphics[width=0.235\textwidth,height=3cm]{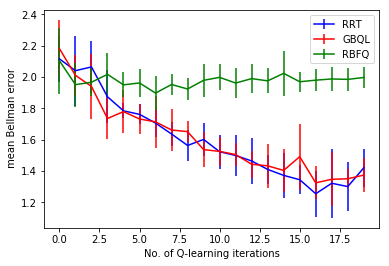}
    \label{mean_bellman_on}
    }
    \subfigure[Logistics]{
    \includegraphics[width=0.235\textwidth,height=3cm]{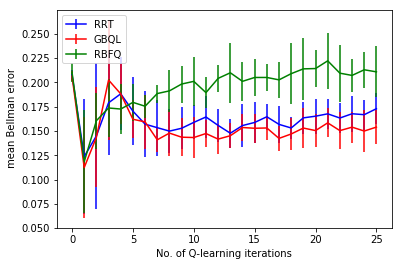}
    \label{mean_bellman_log}
    }
    
  \caption{Mean absolute Bellman training error for different RRL domains }
 \label{mean_bellman_error}
 }
 \end{figure*}
 
 \begin{figure*}[!ht]
    \centering
    {
    \subfigure[Stack]{
    \includegraphics[width=0.235\textwidth,height=3cm]{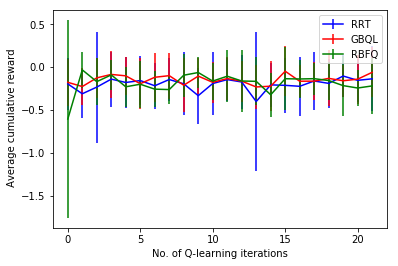} 
    \label{avg_reward_block_sb}}
    \subfigure[Unstack]{
    \includegraphics[width=0.235\textwidth,height=3cm]{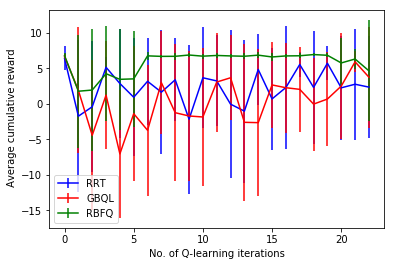} 
    \label{avg_reward_ub}}
    \subfigure[On]{
    \includegraphics[width=0.235\textwidth,height=3cm]{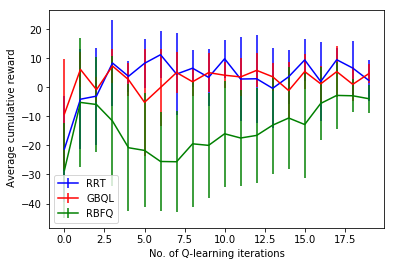} 
    \label{avg_reward_on}}
    \subfigure[Logistics]{
    \includegraphics[width=0.235\textwidth,height=3cm]{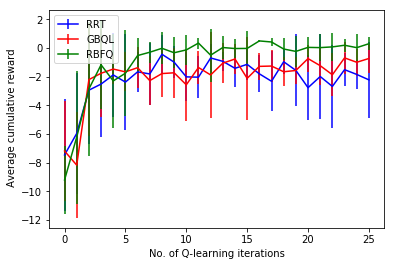} 
    \label{avg_reward_log}}
    
  \caption{Average cumulative reward as a function of the number of Q-learning instances. Higher is better. }
 \label{avg_reward}
 }
 \end{figure*}

We evaluated our proposed $\GBQL$ and $\RBFQ$ algorithms on tasks from the well known Blocks-world domain and the Logistics domain. These domains are rich in structure and are in general considered challenging to many planning/learning algorithms. Unlike several known algorithms, ours is a model-free algorithm that can potentially learn from a small number of trajectories/instances. To test for generalization, we vary the number of objects between training and testing. 

Through the evaluation, we aim to answer the following questions explicitly:
\begin{enumerate}[leftmargin=2.2\parindent]
    \item[\bf Q1:] How does the training error (Bellman residual) vary w.r.t the number of iterations?
    \item[\bf Q2:] How does the test set reward change over time?
    \item[\bf Q3:] How does the performance vary with a differing number of objects during testing (i.e., generalization performance)?
   \item[\bf Q4:] How does $\GBQL$ and $\RBFQ$ compare against each other? 
\end{enumerate}
To this effect, we developed a strong baseline. Specifically, we replace the gradient-boosting of fitted Q-values with a single relational tree (RRT). This is to say that in line $16$ of the $\GBQL$ algorithm, we replace the function call to the boosting algorithm with a function call to a single relational tree learner. Before we explain the experimental baseline, we present the tasks. 

\begin{enumerate}
\item{\bf Blocks World (Stack task):} The blocks world domain~\cite{slaney2001blocks} consists of blocks stacked on top of each other to form towers. The goal of the agent in this task is to stack all the blocks on a single tower. The initial state consists of a random configuration of the blocks on the floor. The state representation for this task consists of predicates like: \texttt{clear}, \texttt{On}, \texttt{heightlessthan}, \texttt{isFloor} etc. The action predicate in this domain is \texttt{move} and the agent learns to move a block from one tower to another to build a single tower. The reward function was set to $+2$ for the goal and $-(1/H_{max}$) for the intermediate steps, where $H_{max}$ refers to the height of the highest tower in a state. The intuition is that configurations closer to the goal state will get a higher reward.
We trained the agent on problems with $3$, $4$ and $5$ number of blocks and tested the policy learnt on problems containing $6$ and $7$ blocks to demonstrate generalization ability.
\item{\bf Blocks World (Unstack task):} The Unstack task is another subproblem in the blocks world domain where the goal of the agent is to unstack all the blocks on the floor. Similar to stack task, the initial state consists of a random configuration of blocks on the floor. The number of predicate is the same and the action predicate for this domain is \texttt{move} where the agent can move a block from one tower to another or to the floor. The reward function is $+10$ for the goal and $-(1-S_i)$ for the intermediate steps where $S_i$ refers to the fraction of blocks on floor in the current state. Reaching the goal configuration is more difficult than the stacking task, hence the reward for the goal state is set to a higher value. The intuition behind the intermediate reward is that the states where a higher fraction of blocks are on the floor are closer to goal and hence should be penalized less. For this domain, we train the agent on $4$, $5$ and $6$ blocks and test the policy learnt on problems containing $7$ blocks. 
\item{\bf{Blocks World (ON task)}:} The goal of the agent is to stack a specific block on top of another block. This problem is comparatively difficult because the optimal policy here is hierarchical in nature and the order of the subgoals needs to be learnt. For example, in order to move a block on top of another block, (1) both the blocks should be clear (2) the upper block should be moved on top of the target block. The state consists of predicates like \texttt{clear}, \texttt{on}, \texttt{sametower}, \texttt{isFloor}, \texttt{goalon} etc. The \texttt{goalon} predicate is used to represent the goal state; \texttt{goalon(b1,b2)} means that the goal is to put block b2 on top of b1. The action predicate is \texttt{move} where the agent can move a block on top of another block or on the floor. The reward is +10 for the goal state. 
For the intermediate states, a small negative reward is provided when a block is moved from a different tower than the goal tower. This is because blocks should be moved from the towers containing the goal blocks in order to make them clear. We train the agent on 4 blocks and test the policy learnt on problems containing 5, 6 and 7 blocks.

\item{\bf Logistics:} This is another classical domain in RRL~\cite{boutilier2001symbolic}. This domain consists of entities like trucks, boxes and cities. The trucks can move from one city to another. The goal of the agent for this task is to unload at least one box in the destination city. The initial state consists of a random configuration of trucks and boxes in all the cities except for the destination city. The state consists of predicates like: \texttt{boxOn}, \texttt{truckIn}, \texttt{boxIn}, \texttt{destination} etc. The actions include the following target predicates: \texttt{load}, \texttt{unload} and \texttt{move}. The reward function is $+1$ for the goal and $-0.2$ for the intermediate steps. We train the agent on problems containing $5$ cities, $3$ trucks and $3$ boxes. We evaluate the learnt policy on problems containing $7$ cities, $3$ trucks and $5$ boxes.
\end {enumerate}

\textbf{Evaluation Methodology}: In every iteration of $\GBQL$ and $\RBFQ$, we sampled $5$ trajectories from a random initial state and test the Q-function learnt by executing the policy greedily on a random $10$ test trajectories. To keep the evaluation fair, we varied the number of objects between train and test trajectories as mentioned earlier for each task. Each of the training and test trajectories is initialized by a random initial state sampled from the initial state distribution of the domain. All the experiments are averaged over $10$ runs. In subsequent Q-learning iterations, we sample trajectories from the simulator using $\epsilon$-greedy policy and the $\epsilon$ is decayed over time by a fixed decay. For every domain, we used $10\%$ of sampling from the previous history also typically known as \textit{experience replay}~\cite{lin1992self} in RL literature. For $\GBQL$, based on the performance on the training trajectories, we chose the number of gradient boosted trees at each iteration to be between $5$ and $10$.

\textbf{Baselines:} As mentioned earlier, we chose a single regression tree (RRT) as the strong baseline. 
This RRT baseline is similar to Dvzeroski et al.'s work~\cite{dvzeroski2001relational} with a key difference that we approximate the Q-values by sampling trajectories from the current policy as opposed to learning from entire experience. Comparison with Deep RL methods is out of scope for this work as the goal is not necessarily to demonstrate the usefulness of gradient-boosting as a function approximator but to demonstrate the ability to incorporate symbolic representations faithfully during learning. Also, as demonstrated in several prior work on Statistical Realtional AI models~\cite{DeRaedt2016}, constructing a flat feature vector from a symbolic structure to train a deep model can lead to loss of information including auto-correlation and linkage~\cite{jensen2002linkage}.

\begin{table}[!ht]
\scalebox{0.95}{
\begin{tabular}{|l|l|l|l|}
\hline
Domain    & RRT        & $\GBQL$        & $\RBFQ$        \\ \hline
Stack     & $0.76 (\pm0.19)$ & $\mathbf{0.82}(\pm0.11)$ & $0.78(\pm0.13)$ \\ 
Unstack   & $0.57(\pm0.32)$ & $\mathbf{0.79}(\pm0.28)$   & $\mathbf{0.79}(\pm0.28)$ \\
On        & $\mathbf{0.32}(\pm0.22) $ & $0.24(\pm0.17)$   & $0.24(\pm0.19)$  \\
Logistics & $0.61(\pm0.21)$  & $0.8(\pm0.12)$    & $\mathbf{0.84}(\pm0.08)$  \\\hline
\end{tabular}}
\caption{$\%$ goals reached on test trajectory by RRT, $\GBQL$ and $\RBFQ$ after executing the learnt Q-function}
\label{table:goals_reached}
\end{table}


\textbf{Results:} To answer \textbf{Q1}, it can be observed from the results reported in Figure~\ref{mean_bellman_error} that the proposed $\GBQL$ and $\RBFQ$ methods clearly demonstrate a decreasing Bellman error as training progresses on the different tasks. 
For the stacking task, the bellman error decreases almost uniformly for both $\GBQL$ and RRT. For $\RBFQ$, the mean absolute bellman error reduces significantly in the beginning but increases with iterations and converges to almost the same point as the other two algorithms.  However, for the logistics domain, it can be seen that $\GBQL$ converges faster than RRT. For the unstack task, it can be seen that the error curve for both $\GBQL$ and RRT follow the same shape and decreases steeply, however, the Bellman error for $\GBQL$ converges faster than \textsc{RRT}. 

For the unstack task, since it is a challenging one, we injected a few expert trajectories instead of fully random exploration in the early iterations. This can be a possible reason for why the Bellman error decreases steeply for this domain. However, induction of expert trajectories does not have any effect on $\RBFQ$. Also, it seems from the results that inductive bias in the form of expert trajectories had more impact on the gradient boosted $\GBQL$ model than RRT as can be seen from the early convergence of $\GBQL$. For the ON task, bellman error for both $\GBQL$ and RRT follows the same shape and converges, however, the bellman error for $\RBFQ$ is quite high and does not reduce with increasing iterations. This allows us to answer \textbf{Q1} affirmatively. 

The average cumulative reward accumulated over the {\em unseen test trajectories with varying number of objects} is shown in Figure~\ref{avg_reward}. The average cumulative reward increases in the early iterations and converges as can be seen from the Figure~\ref{avg_reward_log} especially for logistics domain. Also, in the logistics domain, the average reward for $\RBFQ$ converges much faster than $\GBQL$ and RRT and exhibits less variance. The cumulative reward for unstack task shows a high variance for both $\GBQL$ and RRT. We speculate that the reason for this is the use of {\em reward shaping} which leads to higher variance in this domain. However, in spite of this heuristic being used for the Stacking task, the variance of Stack task is lesser than that of Unstack task. This is because on average, the length of test trajectory to reach the goal is higher for Unstack task than Stack task and thus, the higher variance. For the ON task, $\GBQL$ and RRT converges much faster than $\RBFQ$; however, the cumulative reward collected by $\RBFQ$ increases over time. The above discussions address the question \textbf{Q2} in that the average reward does increase over time.

To address the issue of generalization, we consider the fraction of times that the goal is reached in all the tasks after 20-25 iterations of training in all the domains. Given significant time, all the methods will result in solving most problems. Hence, we computed the percentage of goals reached as the fraction of problems in which the solution was achieved within a certain threshold of the optimal number of steps ($2$ more steps). As can be observed from Table~\ref{table:goals_reached}, $\GBQL$ and $\RBFQ$ achieves a better {\em test set} performance than \textsc{RRT} across in 3 out of 4 tasks. This answers \textbf{Q3} affirmatively in that better generalization is achieved across all the domains by the proposed gradient boosted learning. Finally, it appears that from our experiments, $\RBFQ$ achieves lower Bellman error on some critical states but has a higher error across the states leading to an overall higher error. Our immediate future direction is to investigate deeper the question of whether directly approximating the Bellman error with some guidance can lead to a better average reward. Currently, \textbf{Q4} does not have a definitive answer. The results are split between $\RBFQ$ and $\GBQL$ methods. 

Our initial results demonstrate that both $\RBFQ$ and $\GBQL$ cannot capture effective policies in a hierarchical domain like the ON task. In such a domain, RRT performs better than the two boosted algorithms though the changes are not statistically significant. The Bellman error for $\RBFQ$ increases sometimes like in the logistics domain unlike the other two algorithms where it always decreases. This suggests that since $\RBFQ$ derives it's Q-function as a combination of all the basis function learnt so far, the projection operation on the representable subspace of function is not accurate, hence leading to increase in Bellman error. However, $\RBFQ$ is agnostic to inductive bias in the form of expert (or even noisy) trajectories unlike the other two algorithms as can be seen from Figure~\ref{mean_bellman_ub} of Unstack task  where there is a significant reduction in Bellman error in early iterations for $\GBQL$ and RRT. Also, average cumulative reward on unseen trajectories shows a high variance for Unstack task(Figure~\ref{avg_reward_ub}) for $\GBQL$ and RRT; however $\RBFQ$ is stable and shows lower variance suggesting that $\RBFQ$ is stabler to complex reward functions given we use \textit{reward shaping} in this task to guide the intermediate steps.




\section{Conclusion}
We introduced Relational Approximate Dynamic Programming (ADP) and presented the first set of algorithms $\GBQL$ and $\RBFQ$ that approximates the Q-values or the Bellman error using a non-parameteric gradient-boosting method. The intuition underlying these algorithms is that one can approximate the value function over a set of objects and relations as a set of RRTs learned in a sequential manner. The Bellman operator application step corresponds to the evaluation of these trees for a given state action pair and the projection step corresponds to the learning of these trees using gradient-boosting. Our experiments clearly demonstrate the effectiveness of this approach in factored state space. Most importantly, gradient boosting paves the way to deal jointly
with propositional and relational features; one only has to adapt the gradient regression examples correspondingly. We also demonstrated the generalization ability of the algorithms with our empirical evaluation for intra-domain transfer learning. 

These initial results suggest several interesting avenues for future work. First is evaluating the algorithm on larger problems with hybrid tasks. Second is exploring various sampling strategies in \textsc{GBQL} for picking trajectories from current policy. Third is extending our work to generalized continuous state-action spaces, multi-agent settings and potentially Partially Observable Markov Decision Processes (POMDPs). Next is understanding how rich human inputs can be used to guide the algorithms. Since we are in a symbolic setting, it must be possible for the human to directly provide advice to the system at a high-level. Exploring the use of such knowledge in the context of effective learning remains an interesting direction. Finally, an effective combination of policy gradient and fitted VI methods needs to be explored. 
\section*{Acknowledgements}

SN, SD \& RP gratefully acknowledge the support of NSF grant IIS-1836565. KK acknowledges the support of the Federal Ministry of Economic Affairs and Energy (BMWi) project “Scalable adaptive production systems through AI-based resilience optimization" (SPAICER,  funding reference  01MK20015E), funded within the AI innovation competition “Artificial intelligence as a driver for economically relevant ecosystems“. Any opinions, findings and conclusion or recommendations are those of the authors and do not necessarily reflect the view of the US government.
\bibliographystyle{kr}
\bibliography{reference}

\end{document}